\documentclass[runningheads]{llncs}
\usepackage{float}
\usepackage{amsmath} 
\usepackage[T1]{fontenc}
\usepackage{graphicx}
\usepackage{amsfonts}
\usepackage{multirow}
\usepackage[export]{adjustbox}
\usepackage{xcolor}
\usepackage{booktabs}
\usepackage{arydshln}
\usepackage{subcaption}
\usepackage{hyperref}
\usepackage{acronym}
\usepackage{microtype}
\usepackage{comment}
\begin{document}
\title{
\texorpdfstring{\includegraphics[width=1cm,height=1cm,keepaspectratio]{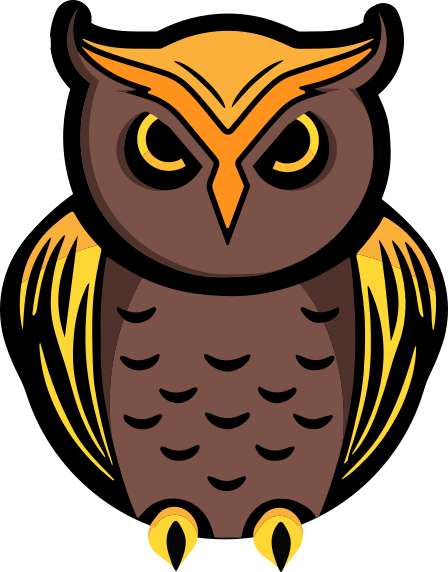}} - ERASMO: Leveraging Large Language Models for Enhanced Clustering Segmentation
}
%
%
\author{Fillipe dos Santos Silva\inst{1,2,3}\and Gabriel Kenzo Kakimoto\inst{1,2,4}\and Julio Cesar dos Reis\inst{1,3} \and Marcelo S. Reis\inst{1,2,3}}
%
%
\institute{$^1$Hub de Inteligência Artificial e Arquiteturas Cognitivas (H.IAAC);\\ $^2$Artificial Intelligence Laboratory (Recod.ai);\\ $^3$Instituto de Computação, Universidade Estadual de Campinas (UNICAMP), Brazil\\
$^4$Faculdade de Engenharia Mecânica, Universidade Estadual de Campinas (UNICAMP), Brazil\\
\email{\{fillipesantos, g234878, msreis, jreis\}@ic.unicamp.br}}
\acrodef{LLM}{Large Language Model}
\acrodef{ARI}{Adjusted Rand Index}
\acrodef{DICE}{Deterministic, Independent-of-Corpus Embeddings}
\acrodef{AHC}{Agglomerative Hierarchical Clustering}
\acrodef{FuzzyCM}{Fuzzy C-Means}
\acrodef{BERT}{Bidirectional Encoder Representations for Transformers}
\acrodef{RAM}{Random Access Memory}
\acrodef{GPU}{Graphics processing unit}
\acrodef{ARI}{Adjusted Rand Index}
\acrodef{HS}{Homogeneity Score}
\acrodef{F1S}{weighted F1-score}
\acrodef{SS}{Silhouette Score}
\acrodef{CHI}{Calinski-Harabasz Index}
\acrodef{DBI}{Davies-Bouldin Index}
\acrodef{BPE}{Byte-Pair-Encodings}
\acrodef{t-SNE}{t-Distributed Stochastic Neighbor Embedding}
\acrodef{RFM}{Recency, Frequency, Monetary}
\acrodef{ML}{Machine Learning}
\acrodef{RAG}{RetrievaFl-Augmented Generation}
\acrodef{MPNet}{ Masked and Permuted Pre-training for Language Understanding }
\maketitle              
\begin{abstract} 
Cluster analysis plays a crucial role in various domains and applications, such as customer segmentation in marketing. These contexts often involve multimodal data, including both tabular and textual datasets, making it challenging to represent hidden patterns for obtaining meaningful clusters. This study introduces ERASMO, a framework designed to fine-tune a pretrained language model on textually encoded tabular data and generate embeddings from the fine-tuned model. ERASMO employs a textual converter to transform tabular data into a textual format, enabling the language model to process and understand the data more effectively. Additionally, ERASMO produces contextually rich and structurally representative embeddings through techniques such as random feature sequence shuffling and number verbalization. Extensive experimental evaluations were conducted using multiple datasets and baseline approaches. Our results demonstrate that ERASMO fully leverages the specific context of each tabular dataset, leading to more precise and nuanced embeddings for accurate clustering. This approach enhances clustering performance by capturing complex relationship patterns within diverse tabular data.
\keywords{Clustering Segmentation  \and Transformer-based Models \and Tabular Data Embeddings.}
\end{abstract}

\section{Introduction}
\label{sec:introduction}

Tabular data is ubiquitous in various fields such as finance, healthcare, and marketing, where it serves as a primary source of information for \ac{ML} tasks~\cite{pitafi2023taxonomy}. 
Despite its widespread use, extracting meaningful insights from tabular data remains a complex challenge, particularly in clustering tasks~\cite{keraghel2024beyond}. These challenges include handling heterogeneous feature types, dealing with high-dimensional spaces, and ensuring meaningful distance metrics. These can significantly impact the effectiveness of clustering algorithms in identifying natural groupings within the data~\cite{petukhova2024text}.

Recent researchers have explored traditional statistical methods and modern deep learning approaches~\cite{petukhova2024text,tissera2024enhancing,viswanathan2023large} to address these challenges. Existing studies have utilized \acp{LLM} like OpenAI's GPT and LLaMA to create embeddings from textual datasets, enhancing data representation and analysis~\cite{keraghel2024beyond,petukhova2024text}. In addition, a method combining \acp{LLM} and \ac{DICE} has been proposed to generate consistent embeddings across datasets, improving segmentation accuracy~\cite{tissera2024enhancing}. However, these approaches often fail to fully leverage the specific context of each tabular dataset, resulting in less nuanced embeddings for precise clustering.

This study originally introduces ERASMO, our proposed framework designed to gen\textbf{ERA}te high-quality embeddings from tabular data using tran\textbf{S}former-based language \textbf{MO}dels. These embeddings excel in clustering analysis, revealing hidden patterns and groupings. Our solution can also be used in \ac{RAG} systems and other tasks to gain deeper insights from context information~\cite{lewis2020retrieval}. ERASMO operates through two stages: \textbf{(1)} fine-tuning a pretrained language model on textually encoded tabular data; and \textbf{(2)} generating embeddings from the fine-tuned model. Using techniques like random feature sequence shuffling and number verbalization, ERASMO produces contextually rich and structurally representative embeddings, outperforming all clustering strategies from the literature based on internal metrics.

Our experimental evaluation used three clustering quality metrics to compare ERASMO with state-of-the-art methods: \ac{SS}, \ac{CHI}, and \ac{DBI}. These metrics comprehensively assess clustering effectiveness results by measuring cohesion, separation, and overall cluster structure.

We extensively evaluated ERASMO on real-world datasets without true labels, including Banking Marketing Targets, E-Commerce Public Dataset by Olist, Yelp reviews, PetFinder.my, and Women's Clothing Reviews. 
These datasets encompass diversified information and present various challenges, rigorously testing ERASMO's clustering capabilities.

This article provides the main contributions as follows:
\begin{itemize}
    \item We introduce ERASMO, a novel framework that leverages transformer-based language models to generate high-quality embeddings from tabular data, enhancing clustering analysis.
    
    \item We demonstrate that our framework significantly improves clustering performance by capturing the complex relationships within tabular data through random feature sequence shuffling and number verbalization techniques.
   
    \item We experimentally achieve state-of-the-art clustering results with ERASMO, showcasing its effectiveness in identifying patterns and groupings within diverse tabular datasets.
   
    \item To the best of our knowledge, ERASMO is the first framework to fully integrate and fine-tune transformer-based language models specifically for generating embeddings from tabular data, resulting in superior clustering outcomes.
\end{itemize}

The remainder of this article is organized as follows. 
Section~\ref{sec:related_works} presents a synthesis of related work. 
Section~\ref{sec:erasmo} details the ERASMO framework. 
Section~\ref{sec:experimental_methodology} outlines our experimental methodology.
Section~\ref{sec:experimental_results} presents the results of our evaluations. 
Section~\ref{sec:discussion} discusses the implications of our findings.
Finally, Section~\ref{sec:conclusion} provides conclusions and directions for future work.
\section{Related Work}
\label{sec:related_works}

Several studies have explored the application of \acp{LLM} to transform tabular data for clustering tasks, demonstrating the potential to enhance user segmentation and data analysis~\cite{keraghel2024beyond,petukhova2024text,tissera2024enhancing,viswanathan2023large,zhang2023clusterllm,zhu2023word}. Zhu \textit{et al.} \cite{zhu2023word} proposed a novel method named Word Embedding of Dimensionality Reduction (WERD) for document clustering. Their approach integrates pre-trained word embeddings with dimensionality reduction techniques. In their work, Sentence-BERT embeds them into high-dimensional vectors after preprocessing documents, which PaCMAP then reduces. Spectral clustering is applied, followed by Non-Negative Matrix Factorization to extract keywords. 

CLUSTERLLM ~\cite{zhang2023clusterllm}, a novel text clustering framework, leverages feedback from \acp{LLM} such as ChatGPT. This method enhances clustering by utilizing \acp{LLM} to refine clustering perspectives and granularity through two stages: a triplet task for fine-tuning embedders based on user preferences and a pairwise task for determining cluster granularity. Extensive experiments on fourteen datasets demonstrated that CLUSTERLLM consistently improves clustering quality and is cost-effective, outperforming traditional clustering methods.
Both WERD \cite{zhu2023word} and CLUSTERLLM \cite{zhang2023clusterllm} presented limitations compared to the ERASMO framework (our proposal). WERD might not fully capture the contextual nuances of each dataset due to its focus on dimensionality reduction techniques. At the same time, CLUSTERLLM's reliance on general-purpose \acp{LLM} for guidance may overlook specific dataset characteristics.

A method demonstrating that \acp{LLM} enables few-short learning applied to clustering tasks was proposed in \cite{viswanathan2023large}. 
Their study showed how \acp{LLM} can perform clustering tasks with minimal labeled data by leveraging their extensive pretraining, significantly reducing the need for large annotated datasets and achieving reasonable clustering performance with few-shot learning. 
Similarly, Tipirneni \textit{et al.} \cite{tipirneni2024context} explored context-aware clustering using \acp{LLM}, highlighting how these models can utilize contextual information to enhance clustering accuracy. Both methods, however, may not fully leverage the dataset-specific nuances as effectively as ERASMO because we employ a fine-tuning step, allowing the model to capture better and utilize dataset-specific details, leading to more accurate and reliable clustering results.

Tissera, Asanka, \& Rajapakse developed ~\cite{tissera2024enhancing} an approach to enhancing customer segmentation using \acp{LLM} and \ac{DICE}. 
Their method combined \acp{LLM} with \ac{DICE} to generate consistent and deterministic embeddings across different datasets, improving segmentation accuracy and robustness. Their approach may not fully leverage the context-specific nuances of each dataset as effectively as ERASMO. Our fine-tuning process proposal allows it to adapt to the unique characteristics of the input data, providing more contextually rich and detailed embeddings that can result in more precise and meaningful clusters.

A comparative analysis of LLM embeddings for effective clustering was explored in ~\cite{keraghel2024beyond}. 
As an extension, the study on text clustering with LLM embeddings~\cite{petukhova2024text} delves deeper, exploring additional models and datasets to demonstrate improvements in text data clustering. While existing approaches effectively capture complex semantic relationships and handle categorical, numerical, and textual data, they lack the fine-tuning specificity of ERASMO, our key originality aspect. ERASMO's tailored embeddings for tabular datasets and integration of feature order permutation provide more precise and contextually relevant clusters, offering superior versatility and robustness in various clustering applications.

\section{ERASMO}
\label{sec:erasmo}

This section introduces ERASMO, our framework that leverages transformer-based language models to generate high-quality embeddings from tabular data. These embeddings are particularly effective for clustering analysis, allowing for identifying patterns and groupings within the data that might not be immediately apparent. 

The process involves two main stages: \textbf{(1)} fine-tuning a pretrained \ac{LLM} on a textually encoded tabular dataset; and \textbf{(2)} utilizing the fine-tuned model to generate embeddings, which are used by a clustering algorithm. These designed stages were inspired by ~\cite{borisov2022language}. Subsection~\ref{subsec:erasmo_phase1} details the fine-tuning phase, whereas  Subsection \ref{subsec:erasmo_phase2} reports on the embedding generation processes.

\subsection{Phase 1: Fine-Tuning}
\label{subsec:erasmo_phase1}

Standard pretrained generative \acp{LLM} expect sequences of words as inputs. Hence, we convert each row of our dataset into a textual representation to apply an \ac{LLM} to tabular data, which can contain categorical, numerical, and textual information.

\begin{definition}[Textual Converter]
Given a tabular dataset with $m$ columns with feature names $f_1, f_2, \ldots, f_m$ and $n$ rows of samples $s_1, \ldots, s_n$, let the entry $v_{i,j}$, $i \in \{1, \ldots, n\}$, $j \in \{1, \ldots, m\}$ represent the value of the $j$-th feature of the $i$-th data point. Taking the feature name and value into account, each sample $s_i$ of the table is transformed into a textual representation $t_i$ using the following subject-predicate-object transformation:
\begin{subequations}
\begin{align}
t_{i,j} &= [f_j, \text{``is''}, v_{i,j}, \text{``,''}], \quad \forall i \in \{1, \ldots, n\}, j \in \{1, \ldots, m\} \\
t_i &= [t_{i,1}, t_{i,2}, \ldots, t_{i,m}], \quad \forall i \in \{1, \ldots, n\},
\end{align}
\end{subequations}
where $t_{i,j}$, the textually encoded feature, is a clause with information about a single value and its corresponding feature name, and $[ \cdot ]$ denotes the concatenation operator.
\end{definition}

By transforming a tabular feature vector into a sequence using the textual subject-predicate-object encoding scheme, pseudo-positional information is artificially introduced into the transformed tabular data sample. However, there is no spatial ordering relationship between features in tabular datasets. We randomly permute the encoded short sentences $t_{i,j}$ of the full textual representation $t_i$ to reconstruct the feature order independence. 


\begin{definition}[Random Feature Sequence Shuffle] Let $t_i, i \in \{1, \ldots, n\}$, be a textual representation. Consider a sequence $k = ( k_1, \ldots, k_m )$ that is a permutation of the sequence of indices $( 1, \ldots, m )$. A random feature sequence shuffle is defined as $t_i(k) = [t_{i,k_1}, t_{i,k_2}, \ldots, t_{i,k_m}]$.
\end{definition}

We fine-tune our generative language model on samples without order dependencies when using shuffled orders of the textually encoded features. Moreover, such permutations are highly beneficial as they allow for arbitrary conditioning in tabular data generation.
In our experiments, we refer to ERASMO\textsubscript{base} as the baseline model, utilizing only the Textual Converter and Random Feature Sequence Shuffle. In addition, there is evidence that verbalizing numerical tokens can enhance effectiveness in specific scenarios \cite{liu2024understanding}. In this sense, we explore this approach, naming it ERASMO\textsubscript{NV}, as follows.


\begin{definition}[Number Verbalizer]
Let $t_i, i \in \{1, \ldots, n\}$, be a textual representation, and $t_{i,j}$ be the set of words of the  $j$-th feature of $t_i$. A number verbalizer is a function $v$ that receives as input a word $w$ of $t_{i,j}$ and is defined as:
\begin{align}
v(w) &= \begin{cases}
        w, \ \mbox{ if $w$ is not numerical},\\
        \mbox{verbalized $w$ otherwise.}
        \end{cases}
\end{align}
\end{definition}
By applying this transformation on every token of every textual representation, we ensure that any numerical information in the text is verbalized. In some NLP tasks, such as clustering with embeddings, sentiment analysis, and text classification, verbalizing numbers can improve the model's understanding of the context and meaning of numerical values, leading to more accurate and meaningful results.  This transformation might not be beneficial in some cases, depending on the specific nature of the data and the task at hand~\cite{liu2024understanding,min2023recent}.

\textbf{Fine-Tuning a Pretrained Auto-Regressive Language Model}: We describe the fine-tuning procedure of a pretrained \ac{LLM} on the encoded tabular data for generation tasks. We suppose a textually encoded tabular dataset $T = \{t_i(k)\}_{i=1, \ldots, n}$ that was transformed into text by the proposed encoding scheme. Let $k$ be a randomly drawn permutation, and $n$ denote the number of rows. Based on user choice, the pipeline can proceed directly to fine-tuning the \ac{LLM} to generate ERASMO\textsubscript{base}, or it can first apply a number verbalizer to convert numerical tokens into their verbal representations before fine-tuning the \ac{LLM} to generate  ERASMO\textsubscript{NV}.

To be processed with an \ac{LLM}, the input sentences $t \in T$ must be encoded into a sequence of tokens from a discrete and finite vocabulary $W$. These tokens can be character, word, or subword encodings such as the \ac{BPE}. Thus, $t \in T$ is represented by a sequence of tokens $(w_1, \ldots, w_j) = \text{TOKENIZE}(t)$ with tokens $w_1, \ldots, w_j \in W$, where $j$ denotes the number of tokens required to describe the character sequence $t$. Commonly, the probability of natural-language sequences is factorized in an auto-regressive manner in \acp{LLM}. It is represented as a product of output probabilities conditioned on previously observed tokens:
\begin{align}
p(t) = p(w_1, \ldots, w_j) = \prod_{k=1}^j p(w_k \mid w_1, \ldots, w_{k-1}).
\end{align}

\begin{figure}
    \centering
    \includegraphics[width=0.54\paperwidth]{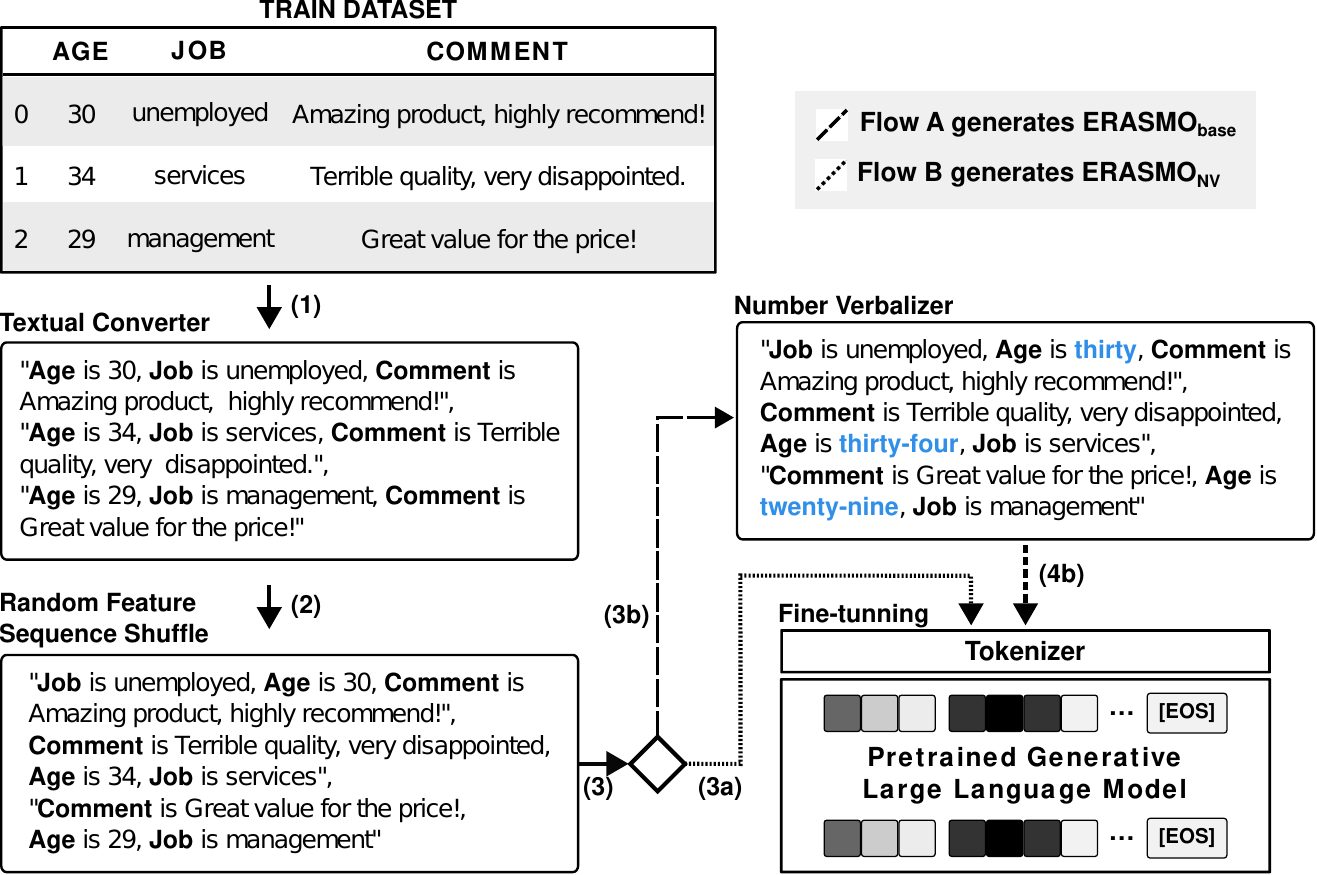}
    \caption{The ERASMO data pipeline for the fine-tuning phase. First, a textual converter step transforms tabular data into meaningful text (1). Next, a random feature order permutation step is applied (2). Then, based on user choice, the pipeline diverges: it can proceed directly to fine-tuning a \ac{LLM} (3a) to generate ERASMO\textsubscript{base}, or apply a number verbalizer (3b) before fine-tuning the \ac{LLM} (4b) to generate  ERASMO\textsubscript{NV}.}
    \label{fig:erasmo_phase1}
\end{figure}

As a result, an end-user can choose any existing generative language model for tabular data modeling and exploit the vast amount of contextual knowledge presented in these models. Fine-tuning enables the model to leverage this contextual information with the feature and category names to enhance the model's capabilities. Figure \ref{fig:erasmo_phase1} presents the pipeline for ERASMO's fine-tuning step.

\subsection{Phase 2: Embedding Generation and Clustering Analysis}
\label{subsec:erasmo_phase2}

We generate embeddings from the model after fine-tuning the \ac{LLM} on the textually encoded tabular dataset. These embeddings capture the contextual relationships and features encoded during the training phase.

We start by feeding the test dataset, transformed into its textual representation, into the fine-tuned \ac{LLM}. The model generates embeddings for each input sequence, providing a high-dimensional representation for each sample. This process ensures that the embeddings preserve the contextual and feature relationships learned during fine-tuning. Depending on the user's choice in the pipeline, the embeddings are generated from either ERASMO\textsubscript{base} or ERASMO\textsubscript{NV} models, reflecting whether the number verbalizer step was applied.

\begin{figure}
    \centering
    \includegraphics[width=0.45\paperwidth]{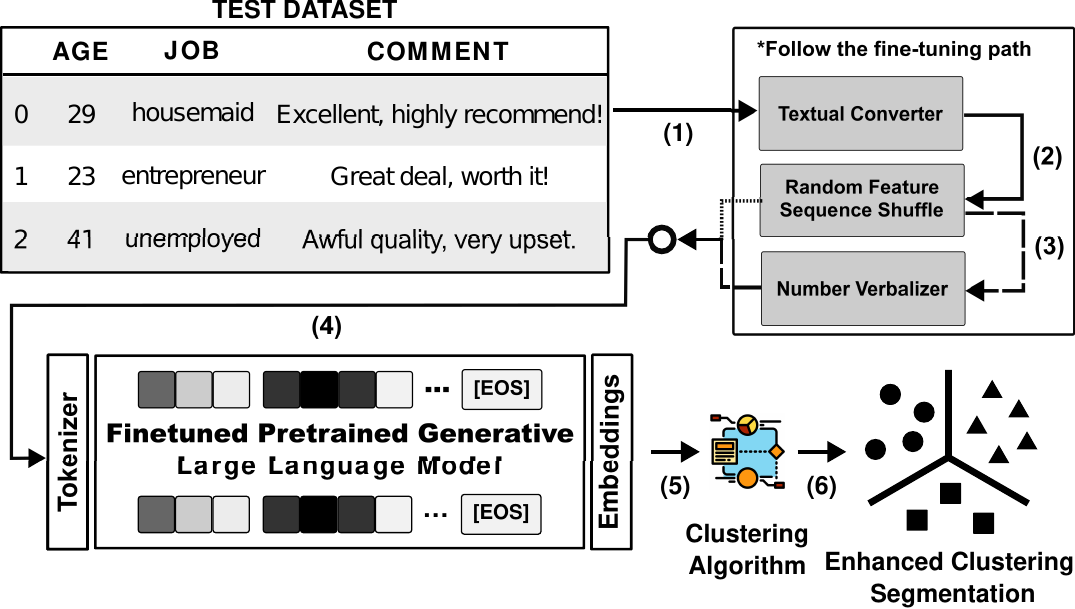}
    \caption{The ERASMO pipeline for generating embeddings and cluster analysis. The input test tabular data is first transformed into text sequences (1). Next, a random feature order permutation step is applied (2). For  ERASMO\textsubscript{NV}, a number verbalizer step follows (3) before processing by the fine-tuned \ac{LLM} to generate embeddings (4). For ERASMO\textsubscript{base}, the pipeline goes directly from step (2) to step (4). The embeddings are subsequently used for clustering analysis.}
    \label{fig:erasmo_phase2}
\end{figure}

To generate these embeddings, the input sentences $t \in T_{test}$ are encoded into sequences of tokens and processed by the fine-tuned \ac{LLM}. The embeddings are obtained from the final hidden states of the model, resulting in rich and informative representations of the data. These embeddings can then be utilized for various downstream tasks, including clustering analysis, to gain deeper insights into the data structure (cf. Figure~\ref{fig:erasmo_phase2}).

\section{Experimental Methodology}
\label{sec:experimental_methodology}

Our experiments evaluated the quality assessment for the best-performing clustering algorithms for each dataset and approach (model) combination (cf. Table~\ref{tab:clustering_quality_assessment} for the obtained results). Subsection~\ref{subsec:datasets} describes the datasets used for training and testing. Subsection~\ref{subsec:Clustering algorithms} presents an overview of the clustering algorithms. Subsection~\ref{subsec:baselines} describes the approaches used for comparison as baselines in our experiments. Subsection~\ref{subsec:Evaluation Metrics} reports on the evaluation metrics. Subsection~\ref{subsec:Implementation Details} presents the implementation details. Each experimental setup for a given dataset assessed the different language models fine-tuned and pretrained. Each model considers the several clustering algorithms and their configuration.  

\subsection{Datasets}
\label{subsec:datasets}
We selected a diversified set of datasets to encompass a variety of challenges related to text categorization and clustering, and we used them to evaluate text clustering algorithms.

\begin{itemize}
    \item \textbf{Banking Marketing Targets:} Composed of data from direct marketing campaigns of a banking institution, which includes client attributes like age and job, along with the response to the campaign~\cite{misc_bank_marketing_222}.

    \item \textbf{E-Commerce Public Dataset by Olist:} A Brazilian e-commerce dataset with over 100,000 orders from 2016 to 2018 across multiple marketplaces~\cite{olist2023}. It includes 72,794 training and 18,199 testing samples. The \ac{RFM} model was used for customer segmentation, as described in~\cite{tissera2024enhancing}.

    \item \textbf{Yelp:} Comprises reviews from Yelp businesses, including text reviews, star ratings, and business attributes, offering a rich resource for sentiment analysis and review classification tasks~\cite{Yelp-Dataset}.
    
    \item \textbf{PetFinder.my:} Features adoption records from the PetFinder.my website, encompassing various pet attributes, descriptions, and adoption status, valuable for text classification and clustering related to animal welfare~\cite{kaggle-petfinder-adoption-prediction}.
    
    \item \textbf{Women Clothing Reviews:} Contains reviews of women's clothing, with each review detailing text feedback, ratings, and customer information, suitable for sentiment analysis and recommendation system research~\cite{Womens-e-commerce-clothing-reviews}.
    
\end{itemize}

Each unlabeled dataset was processed through the proposed pipeline, which involves training a pretrained \ac{LLM}. This approach ensures that the clustering algorithms can perform optimally across diverse textual inputs, enhancing their ability to effectively identify and group related data points.

\subsection{Clustering Algorithms}
\label{subsec:Clustering algorithms}

The clustering algorithms chosen are well-suited for handling complex patterns in structured and textual data, ensuring efficient categorization.

We used the \textit{k}-means algorithm for its simplicity and effectiveness with large datasets and \textit{k}-means++ for its strategic centroid initialization to enhance clustering efficiency and quality~\cite{ourabah2023large}. Unlike \textit{k}-means, which assigns each data point to a single cluster, \ac{FuzzyCM} employs a probabilistic membership approach, effectively capturing the nuances and polysemy typical of textual data.
We used \ac{AHC} to uncover hierarchical structures and spectral clustering for its proficiency in recognizing clusters based on the data's graph structure, effectively identifying non-convex shapes.

For \textit{k}-means, the parameters were: initialization method set to random, number of initializations (\textit{ninit}) set to 10, and the random seed set to 0. The \textit{k}-means++ algorithm utilized \textit{k}-means++ for initialization, \textit{ninit} was 1, and the seed was 0. Agglomerative Hierarchical Clustering (AHC) employed the Euclidean metric with Ward linkage. For Fuzzy C-means (FuzzyCM), no initialization method was specified, the fuzziness parameter (\textit{m}) was 2, error tolerance was set to 0.005, and the maximum number of iterations (\textit{maxiter}) was 1000. Spectral clustering used 'discretize' for label assignment and a random seed of 10.

Implementations for these algorithms were sourced from the \textit{scikit-learn library} \cite{pedregosa2011scikit}, except for \ac{FuzzyCM}, which used the \textit{scikit-fuzzy} package~\cite{warner2019scikitfuzzy}. For \textit{k}-means and \textit{k}-means++, \textit{init} specifies the initial cluster centroid method, \textit{ninit} indicates the number of algorithm runs with different seeds, and \textit{seed} sets the random number for centroid initialization. 
In \ac{AHC}, \textit{metric} is the metric used for linkage computation, and \textit{linkage} is the criterion measuring the distance between observation sets. 
We used Euclidean distance to measure point similarities and a nearest centroid approach to associate clusters.

For \ac{FuzzyCM}, \textit{init} is the initial fuzzy c-partitioned matrix (random if \textit{None}), \textit{m} is the fuzziness degree, \textit{error} is the stopping criterion, and \textit{maxiter} is the iteration limit. 
In Spectral clustering, \textit{assign\_labels} specifies the labeling strategy in the embedding space, and \textit{seed} is the pseudorandom number for initializing the eigenvector decomposition.
For all datasets, the number of clusters (\( k \)) was determined using the silhouette score to optimize cluster cohesion and separation.

\subsection{Approaches (Baselines)}
\label{subsec:baselines}

We utilized various embedding techniques from state-of-the-art \acp{LLM}, including OpenAI, Falcon, Llama 2, GPT-2 Medium, and an MPNet-based model, each enhancing text representation by capturing contextual nuances.

For the MPNet-based model, we used \textbf{sentence-transformers/all-mpnet-base-v2} (MPNet-v2)~\cite{song2020mpnet}. For the OpenAI model, we utilized \textbf{text-embedding-3-large}, and for the Falcon model, we used \textbf{tiiuae/falcon-7b}~\cite{almazrouei2023falcon}. Additionally, we employed \textbf{Llama-2-7b-chat-hf}~\cite{touvron2023llama} for chat applications and \textbf{gpt2-medium}~\cite{radford2019language} for faster textual representations. We used the embeddings from the last layer of all models for the most contextually rich text representations.

We integrated an additional baseline from a recent study that explores customer segmentation using \acp{LLM} combined with \ac{DICE}~\cite{tissera2024enhancing}. They used the \textbf{paraphrase-multilingual-mpnet-base-v2} model from Sentence Transformers to generate 768-dimensional sentence embeddings. This model, based on the MPNet architecture, has about 278 million parameters and is designed for clustering and semantic search.

\subsection{Evaluation Metrics}
\label{subsec:Evaluation Metrics}

We use a set of metrics to evaluate our proposed framework and baselines thoroughly. 
Specifically, we employed the \ac{SS}, \ac{CHI}, and \ac{DBI} metrics to assess the cohesion, compactness, and separation of clusters, ensuring a robust analysis of their structural integrity. 

The \ac{SS} metric, which assesses the separation and cohesion of clusters, is calculated for each data point \(i\) as:
\begin{displaymath}
\resizebox{!}{0.9\baselineskip}{$
s(i) = \frac{b(i) - a(i)}{\max\{a(i), b(i)\}},
$}
\end{displaymath}
where \(a(i)\) measures the average intra-cluster distance, and \(b(i)\) is the minimum inter-cluster distance for the point \(i\).

\ac{CHI} measures the ratio of between-cluster dispersion to within-cluster dispersion, providing insights into the overall clustering structure. The index is formulated as:
\begin{displaymath}
\resizebox{!}{0.8\baselineskip}{$
CHI = \frac{\text{Tr}(B_k) / (k-1)}{\text{Tr}(W_k) / (N-k)},
$}
\end{displaymath}
where \(\text{Tr}(B_k)\) is the trace of the between-group dispersion matrix, and \(\text{Tr}(W_k)\) the trace of the within-group dispersion matrix, thus evaluating both the separation and compactness of the clusters.

The \ac{DBI} evaluates the average similarity ratio of each cluster with its most similar one, offering a measure of cluster separation. The index is calculated as:
\begin{displaymath}
\resizebox{!}{0.9\baselineskip}{$
DBI = \frac{1}{k} \sum_{i=1}^{k} \max_{j \neq i} \left( \frac{\sigma_i + \sigma_j}{d_{ij}} \right),
$}
\end{displaymath}
where \(\sigma_i\) and \(\sigma_j\) represent the average distance of all elements in clusters \(i\) and \(j\) to their respective centroids, and \(d_{ij}\) is the distance between centroids of clusters \(i\) and \(j\). Lower values of \ac{DBI} indicate better cluster separation.

\subsection{Our implemented Setup}
\label{subsec:Implementation Details}

We compare the described baselines (cf. Subsection \ref{subsec:baselines}) using a pretrained transformer-decoder LLM model, $GPT-2$ medium~\cite{radford2019language}, which has $355$ million trainable parameters, 24 layers, 16 attention heads, an embedding size of 1024, and a context size of 1024. 
To facilitate this comparison, we convert the tabular dataset into text for all baselines, applying the random feature sequence shuffling function and comparing the results with ERASMO\textsubscript{base} and ERASMO\textsubscript{NV}. 
Both models were trained with a batch size of 8 over 60 epochs. We applied a dropout rate of 0.1 and utilized 500 warmup steps. The models incorporated a weight decay of 0.01 and used the Adam optimizer with $\epsilon$ set to 1e-8 and $\beta$ values of [0.7, 0.9]. The initial learning rate was set to 5e-5, with a schedule starting at 1e-8, ranging from a minimum of 1e-5 to a maximum of 4e-5.
The model was developed using PyTorch~\footnote{\href{https://pytorch.org/}{pytorch.org}.} and is made available at a GitHub repository~\footnote{\href{https://github.com/fsant0s/ERASMO}{ERASMO - GitHub}.}. It ran on a system equipped with five NVIDIA RTX A6000, each having 48 GB of \ac{RAM}.

\section{Experimental Results}
\label{sec:experimental_results}

We present key results obtained organized by datasets. Table~\ref{tab:clustering_quality_assessment} presents the \ac{SS}, \ac{CHI}, and \ac{DBI} metrics results for the test dataset for all evaluated approaches; values in bold indicate the best outcomes. The best algorithm was determined by choosing the algorithm with the highest~\ac{SS}.

\textbf{Banking.} In the Banking dataset, the ERASMO\textsubscript{base} strategy outperformed all other strategies, achieving the highest ~\ac{SS} of $0.75$, the highest \ac{CHI} of $12,038.44$, and the lowest \ac{DBI} of $0.37$. This indicates well-defined and compact clusters. ERASMO\textsubscript{NV} also performed strongly with an \ac{SS} of $0.71$ and \ac{CHI} of $7,570.38$, though it had a slightly higher \ac{DBI} of $0.43$ compared to ERASMO\textsubscript{base}. Other strategies like MPNet-v2 and Falcon showed moderate performance with \ac{SS} values of $0.27$ and $0.23$, respectively, while OpenAI had the lowest \ac{SS} at $0.11$ and the highest \ac{DBI} at $2.73$, indicating poor clustering.

\textbf{Olist.} For the Olist dataset, ERASMO\textsubscript{NV} achieved the highest \ac{SS} of $0.77$, indicating the best clustering quality. It reached the highest \ac{CHI} of $62,036.87$ and a low \ac{DBI} of $0.32$. ERASMO\textsubscript{base} followed closely with an \ac{SS} of $0.75$ and a \ac{CHI} of $54236.31$, along with the lowest \ac{DBI} of $0.30$. Other strategies like LLaMA-2 and Falcon performed reasonably well, with \ac{SS} values of $0.71$ and $0.66$, respectively. However, OpenAI and MPNet-v2 showed lower \ac{SS} values, with OpenAI achieving an \ac{SS} of $0.19$ and MPNet-v2 an \ac{SS} of $0.24$.

\textbf{Yelp.} In the Yelp dataset, ERASMO\textsubscript{NV} and ERASMO\textsubscript{base} both demonstrated superior performance, with \ac{SS} values of $0.79$ and $0.78$, respectively. ERASMO\textsubscript{NV} also achieved the highest \ac{CHI} of $8,410.94$ and tied with ERASMO\textsubscript{base} for the lowest \ac{DBI} of $0.28$. Other strategies, such as GPT2 Medium and LLaMA-2, showed moderate clustering performance with \ac{SS} values of $0.39$ and $0.29$, respectively. OpenAI, with an \ac{SS} of $0.07$, and MPNet-v2, with an \ac{SS} of $0.23$, indicated less effective clustering.

\textbf{PetFinder.my.} In the PetFinder.my dataset, ERASMO\textsubscript{NV} slightly outperformed ERASMO\textsubscript{base} with an \ac{SS} of $0.73$ compared to $0.72$. ERASMO\textsubscript{base} had the highest \ac{CHI} of $3,351.95$ and a low \ac{DBI} of $0.40$, while ERASMO\textsubscript{NV} had a \ac{CHI} of $3,063.55$ and the lowest \ac{DBI} of $0.34$. Other strategies, such as GPT2 Medium and Falcon, showed moderate results with \ac{SS} values of $0.55$ and $0.20$, respectively. MPNet-v2 had the lowest \ac{SS} of $0.14$, indicating poor clustering performance.

\textbf{Clothings.} For the Clothings dataset, ERASMO\textsubscript{base} achieved the highest \ac{SS} of $0.72$ and the highest \ac{CHI} of $6,208.52$, along with the lowest \ac{DBI} of $0.39$. ERASMO\textsubscript{NV} also performed well with an \ac{SS} of $0.71$ and a \ac{CHI} of $5,916.57$, matching the lowest \ac{DBI} of $0.39$. Other strategies like GPT2 Medium and LLaMA-2 showed moderate performance, with \ac{SS} values of $0.52$ and $0.37$, respectively. OpenAI and MPNet-v2 had the lowest \ac{SS} values of $0.07$ and $0.12$, respectively, indicating poor clustering quality.

\begin{table*}
\caption{
This table shows the clustering quality assessment results for the top-performing algorithms across each dataset and approach combination. The best algorithm was chosen based on the highest SS. We provide the optimal number of clusters (\textit{k}) and the results for \ac{SS}, \ac{CHI}, and \ac{DBI}. Bold values indicate the best results for each metric.
}
\label{tab:clustering_quality_assessment}
\centering
\renewcommand\arraystretch{1}
\begin{adjustbox}{width=\textwidth,center}
\setlength{\tabcolsep}{3pt} 
\begin{tabular}{llcccrc}
\hline
Dataset                  & Approach    & Best alg. & Best \textit{k} & \ac{SS} & CHI & DBI \\ \hline
\multirow{8}{*}{Banking} & MPNet-v2        & \textit{k}-means & 2 & 0.27 & 1,981.53 & 1.46 \\
                         & OpenAI      & \textit{k}-means & 9 & 0.11 & 212.96 & 2.73 \\
                         & LLaMA-2     & \textit{k}-means++ & 8 & 0.22 & 593.66 & 1.66 \\
                         & Falcon      & \textit{k}-means & 2 & 0.23 & 1,776.67 & 1.56 \\
                         & GPT2 Medium & \textit{k}-means & 2 & 0.40 & 4,764.08 & 0.90 \\
                         & PMV2 + DICE & \textit{k}-means & 2 & 0.31 & 2,389.25 & 1.33 \\ \cdashline{2-7}
                         & ERASMO\textsubscript{base} & \textit{k}-means & 2 & \textbf{0.75} & \textbf{12,038.44} & \textbf{0.37} \\
                         & ERASMO\textsubscript{NV}  &  AHC & 2 & 0.71 & 7,570.38 & 0.43 \\
\midrule
\multirow{8}{*}{Olist}   & MPNet-v2        & \textit{k}-means & 2 & 0.24 & 5,927.77 & 1.59 \\
                         & OpenAI      & \textit{k}-means& 3 & 0.19 & 3,946.14 & 1.83 \\
                         & LLaMA-2     & \textit{k}-means & 4 & 0.71 & 43,306.34 & 0.45 \\
                         & Falcon      & \textit{k}-means & 6 & 0.66 & 45,512.63& 0.55 \\
                         & GPT2 Medium & \textit{k}-means & 2 & 0.48 & 26,471.54 & 0.75\\
                         & PMV2 + DICE & SC & 2 & 0.61 & 27,578.16 & 0.67 \\ \cdashline{2-7}
                         & ERASMO\textsubscript{base} & SC & 2 & 0.75 & 54,236.31 & \textbf{0.30} \\
                         & ERASMO\textsubscript{NV} & \textit{k}-means & 2 &  \textbf{0.77} & \textbf{62,036.87} & 0.32 \\   
\midrule
\multirow{8}{*}{Yelp}   & MPNet-v2        & AHC & 2 & 0.23 & 36.61 & 2.25 \\
                         & OpenAI      & AHC & 2 & 0.07 & 38.78 & 3.86 \\
                         & LLaMA-2     & \textit{k}-means & 10 &0.29 & 445.87 & 1.42 \\
                         & Falcon      & \textit{k}-means++ & 14 & 0.32 & 442.83 & 1.22 \\
                         & GPT2 Medium & \textit{k}-means & 2 & 0.39  & 1,898.25 & 1.00 \\
                         & PMV2 + DICE & AHC & 2 & 0.53 & 32.86 & 1.08 \\ \cdashline{2-7}
                         & ERASMO\textsubscript{base} & SC & 2 & 0.78 & 7,702.89 & \textbf{0.28}\\
                         & ERASMO\textsubscript{NV} & AHC & 2 & \textbf{0.79} & \textbf{8,410.94} & \textbf{0.28} \\   
\midrule
\multirow{8}{*}{PetFinder.my}   & MPNet-v2 & \textit{k}-means & 2 & 0.14 & 236.58 & 2.28 \\
                         & OpenAI      & AHC & 2 & 0.16 & 3.29 & 1.75 \\
                         & LLaMA-2     & \textit{k}-means++ & 17 & 0.35 & 179.00 & 1.38 \\
                         & Falcon      & \textit{k}-means++ & 2 & 0.20 & 397.65 & 1.86 \\
                         & GPT2 Medium & AHC & 2 & 0.55 & 636.2 & 0.70 \\
                         & PMV2 + DICE & \textit{k}-means & 5 & 0.18 & 242.83 & 1.85 \\ \cdashline{2-7}
                         & ERASMO\textsubscript{base} & \textit{k}-means & 2 & 0.72 & \textbf{3,351.95} & 0.40 \\
                         & ERASMO\textsubscript{NV} & AHC & 2 & \textbf{0.73} & 3,063.55 & \textbf{0.34} \\   
\midrule
\multirow{8}{*}{Clothings}   & MPNet-v2 & \textit{k}-means & 3 & 0.12 & 195.01 & 2.47\\
                         & OpenAI      & SC & 5 & 0.07 & 70.17 & 2.90 \\
                         & LLaMA-2     & \textit{k}-means++ & 9 &  0.37 & 435.45 & 1.53 \\
                         & Falcon      & \textit{k}-means & 12 & 0.24 & 326.64 & 1.54 \\
                         & GPT2 Medium & \textit{k}-means & 2 &  0.52 & 3,541.87 & 0.68\\
                         & PMV2 + DICE & AHC & 2 & 0.17 & 72.74 & 1.97\\ \cdashline{2-7}
                         & ERASMO\textsubscript{base} & \textit{k}-means & 2 & \textbf{0.72} & \textbf{6,208.52} & \textbf{0.39} \\
                         & ERASMO\textsubscript{NV} & \textit{k}-means++ & 2 & 0.71 & 5,916.57& \textbf{0.39} \\   
\bottomrule
\end{tabular}
\end{adjustbox}
\end{table*}

Figure~\ref{fig:tnse} shows a 2D \ac{t-SNE} visualization for the Yelp dataset across all strategies. The Yelp dataset was chosen for its rich and diverse user reviews, making it ideal for clustering evaluation. Our models, ERASMO\textsubscript{base} and ERASMO\textsubscript{NV}, display distinct and well-separated clusters, reflecting their lowest \ac{DBI} scores. This highlights the superior clustering performance of our models.

\newcommand\tfigsct{0.24}
\begin{figure}[!htb]
    \centering
    \begin{subfigure}[]{\tfigsct\textwidth}
        \includegraphics[width=\textwidth]{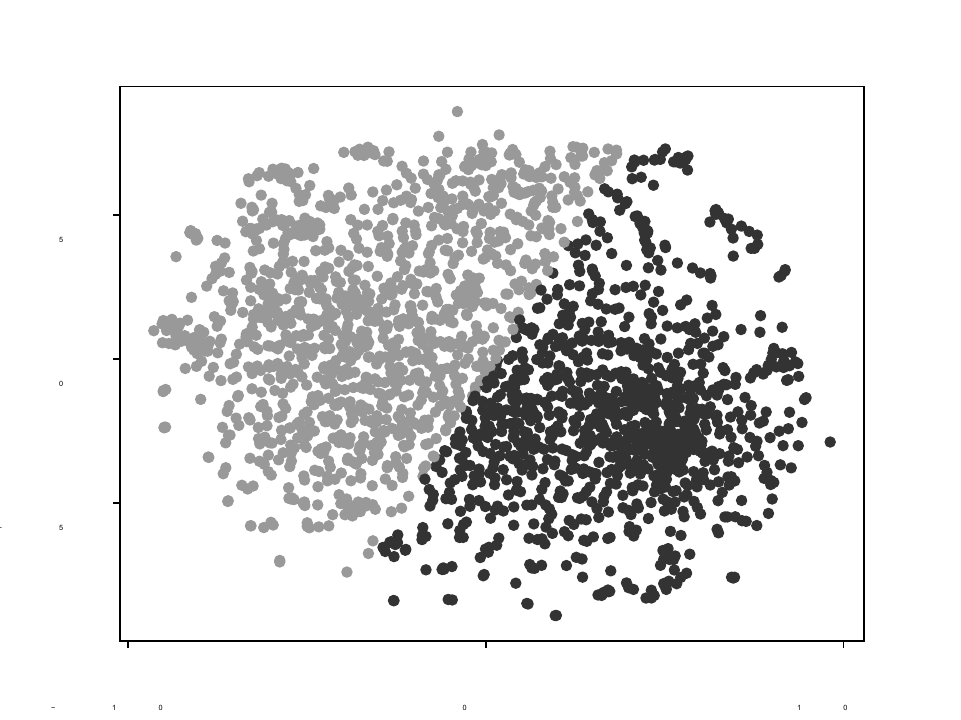}
        \caption{MPNet-v2}
        \label{fig:tsne_bert}
    \end{subfigure}
    \begin{subfigure}[]{\tfigsct\textwidth}
        \includegraphics[width=\textwidth]{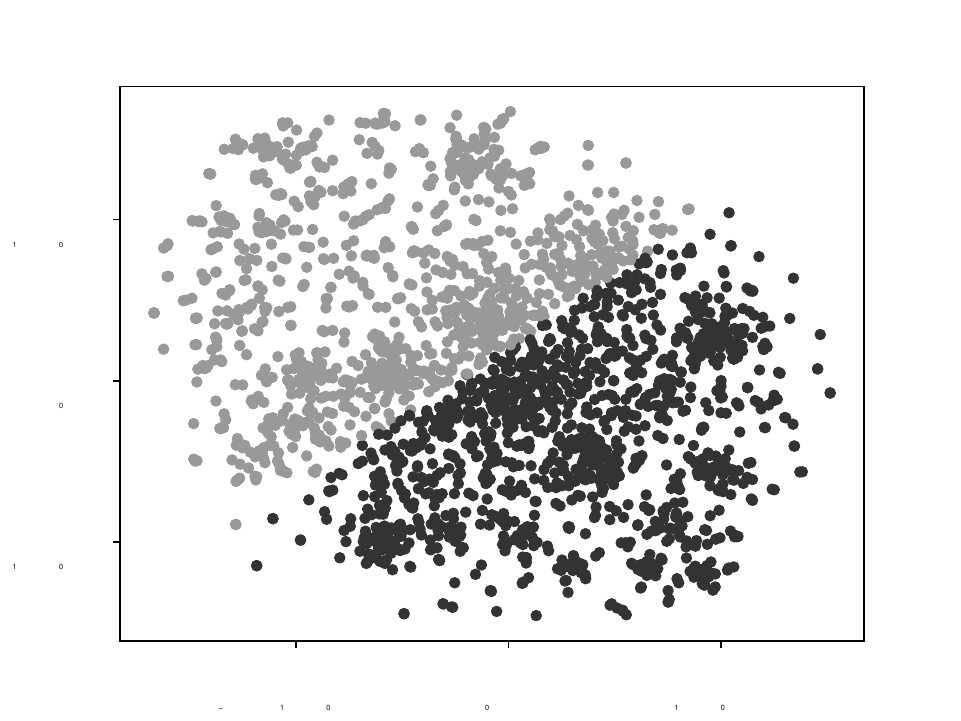}
        \caption{OpenAI}
        \label{fig:tsne_openai}
    \end{subfigure} 
    \begin{subfigure}[]{\tfigsct\textwidth}
        \includegraphics[width=\textwidth]{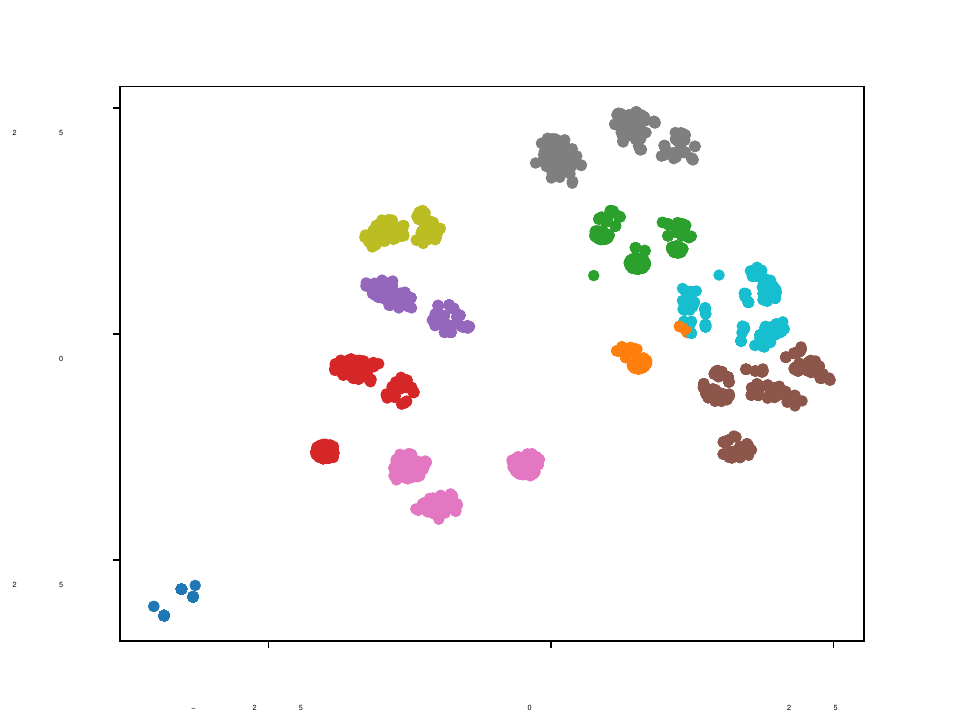}
        \caption{LLaMA-2}
        \label{fig:tsne_llama}
    \end{subfigure}
    \begin{subfigure}[]{\tfigsct\textwidth}
        \includegraphics[width=\textwidth]{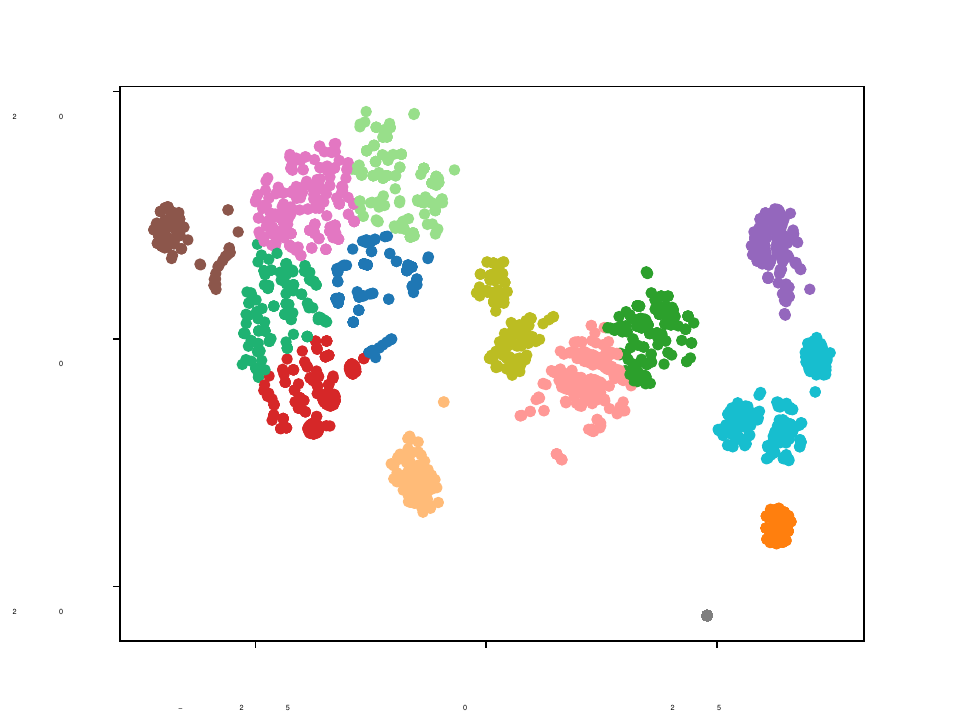}
        \caption{Falcon}
        \label{fig:tsne_falcon}
    \end{subfigure}\\

    \begin{subfigure}[]{\tfigsct\textwidth}
        \includegraphics[width=\textwidth]{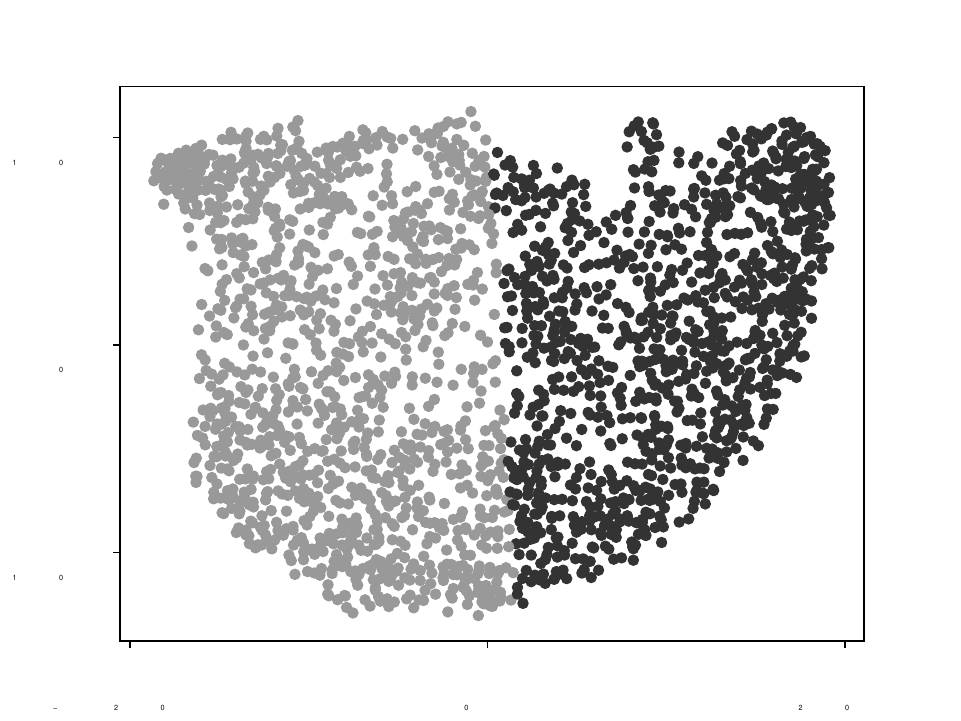}
        \caption{GPT2 Medium}
        \label{fig:tsne_gpt2_medium}
    \end{subfigure}
    \begin{subfigure}[]{\tfigsct\textwidth}
        \includegraphics[width=\textwidth]{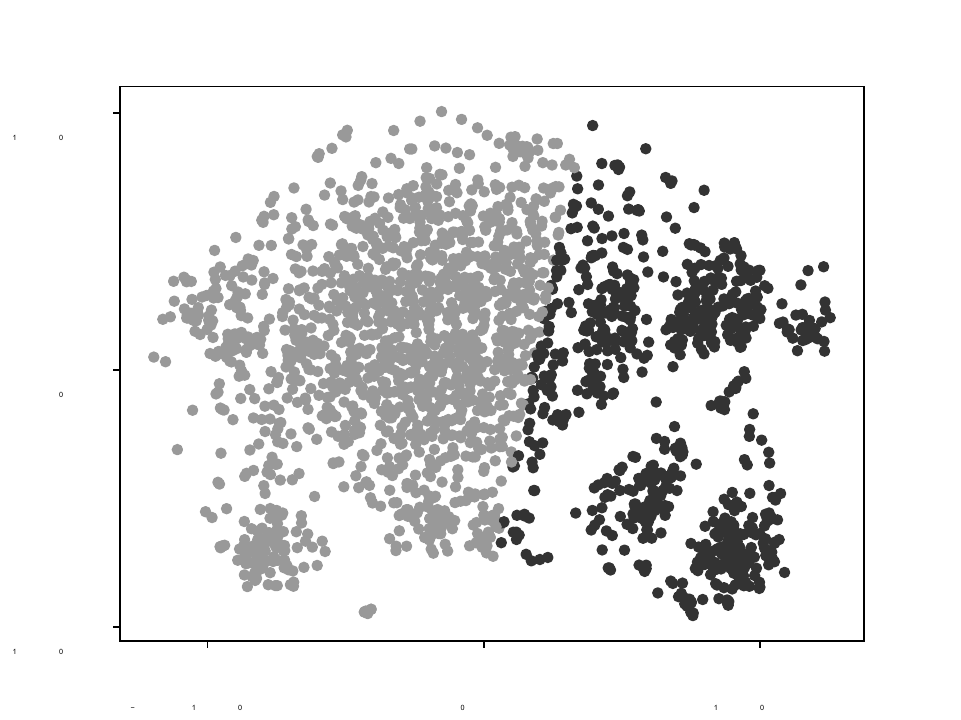}
        \caption{PMV2 + DICE}
        \label{fig:tsne_dice}
    \end{subfigure} 
    \begin{subfigure}[]{\tfigsct\textwidth}
        \includegraphics[width=\textwidth]{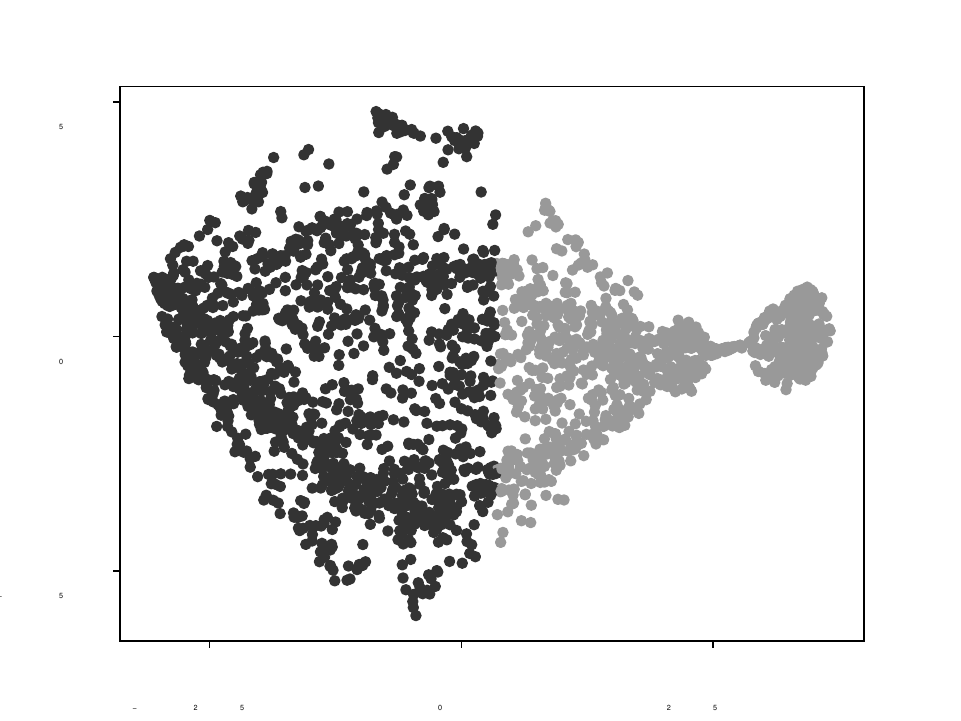}
        \caption{ERASMO\textsubscript{base}}
        \label{fig:tsne_erasmo_false}
    \end{subfigure}
    \begin{subfigure}[]{\tfigsct\textwidth}
        \includegraphics[width=\textwidth]{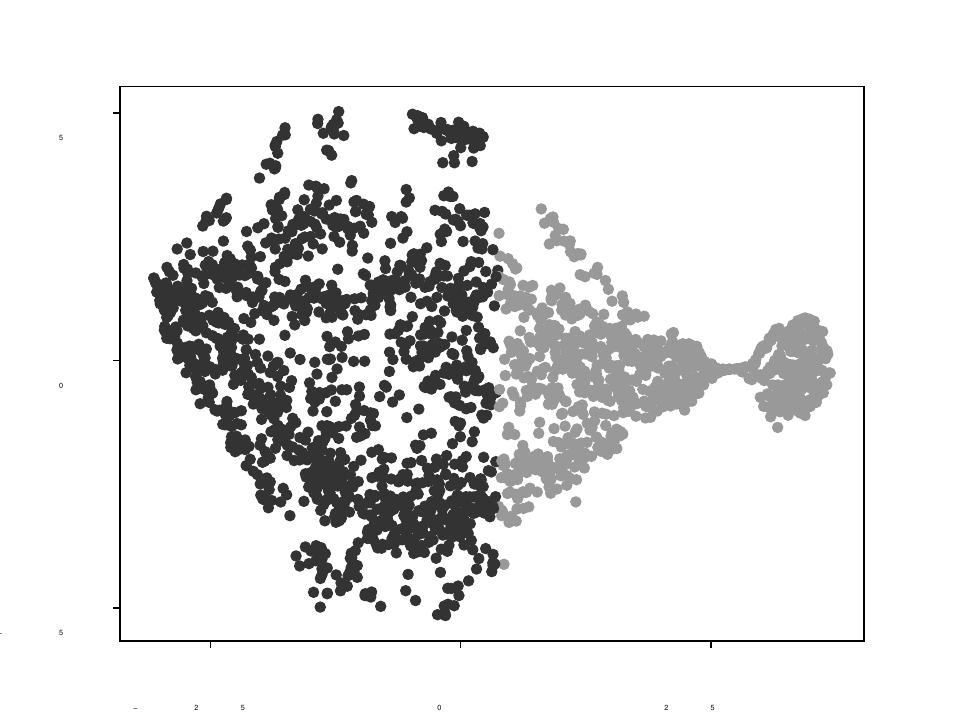}
        \caption{ERASMO\textsubscript{NV}}
        \label{fig:tsne_erasmo_true}
    \end{subfigure}
    \caption{\ac{t-SNE} visualization of embedding representations on the Yelp dataset for different models
    : (a) MPNet-v2, (b) OpenAI, (c) LLaMA-2, (d) Falcon, (e) GPT2 Medium, (f) PMV2 + DICE, (g) ERASMO\textsubscript{base}, and (h) ERASMO\textsubscript{NV}.}
    \label{fig:tnse}
\end{figure}

\section{Discussion}
\label{sec:discussion}
The \ac{SS} results consistently highlight the superior clustering effectiveness of the ERASMO\textsubscript{base} and ERASMO\textsubscript{NV} strategies across all datasets. ERASMO\textsubscript{base} achieved the highest \ac{SS} in the Banking and Clothing datasets, while ERASMO\textsubscript{NV} led in the Olist, Yelp, and PetFinder.my datasets. This indicates that both strategies help to form well-defined clusters, with ERASMO\textsubscript{base} having a slight edge in some datasets. The higher \ac{SS} values for these models suggest they are better at creating distinct clusters than other strategies like MPNet-v2, OpenAI, and Falcon, which showed significantly lower \ac{SS} values, indicating poorer representation relevance for clustering quality.

The \ac{DBI} results further support the effectiveness of ERASMO\textsubscript{base} and ERASMO\textsubscript{NV}. Both strategies consistently achieved the lowest \ac{DBI} values, particularly excelling in the Banking, Olist, and Yelp datasets. A lower \ac{DBI} indicates more compact and well-separated clusters, affirming that our proposed models form tight and distinct clusters. ERASMO\textsubscript{base} generally showed slightly better \ac{DBI} scores, suggesting it produces more compact clusters than ERASMO\textsubscript{NV}. In contrast, approaches like OpenAI and MPNet-v2 had significantly higher \ac{DBI} values, reflecting less adequate cluster compactness and separation.

The \ac{CHI} results corroborate the trends observed in \ac{SS} and DBI. ERASMO\textsubscript{base} and ERASMO\textsubscript{NV} achieved the highest \ac{CHI} scores across most datasets, with ERASMO\textsubscript{base} leading in the Banking and Clothing datasets, and ERASMO\textsubscript{NV} in the Olist and Yelp datasets. High \ac{CHI} scores indicate that our proposed models create clusters that are not only well-separated but also highly dense. These superior results across multiple metrics underscores the robustness and effectiveness of the ERASMO approach. Other models like MPNet-v2, OpenAI, and Falcon demonstrated lower \ac{CHI} scores, indicating less effective clustering effectiveness.

ERASMO has limitations, including reliance on high-quality data, challenges with ambiguous datasets, and significant computational demands for fine-tuning. Random sequence shuffling can affect reproducibility, and verbalizing numerical tokens does not consistently improve outcomes, warranting further investigation.
Furthermore, applying traditional metrics such as SS, CHI, and DBI to analyze clusters from various representation spaces appears as an open research challenge yet to be resolved in the literature. While these metrics are designed for uniform representation spaces, they are frequently applied to different representation spaces, as evidenced by previous studies~\cite{keraghel2024beyond,petukhova2024text}. 
We recognize these limitations and stress the relevance of investigating and developing further refined metrics capable of accurately assessing embeddings from varied dimensional spaces.

Despite these challenges, ERASMO significantly advances clustering techniques by enhancing data representation. This innovation sets the stage for the potential standardization of metrics and improves clustering quality. ERASMO stands as a crucial development in pursuing more robust clustering strategies.
\section{Conclusion}
\label{sec:conclusion}

Exploring structured and textual data simultaneously for clustering analysis is a challenging problem. This study presented and evaluated the ERASMO\textsubscript{base} and ERASMO\textsubscript{NV} clustering approaches, demonstrating their superior empirical effectiveness across multiple datasets. The results, based on \ac{SS}, \ac{DBI}, and \ac{CHI}, consistently revealed that our proposed models help to create well-defined, compact, and dense clusters, outperforming all strategies. Despite their high computational demands and dependency on data quality, we found the ERASMO models suited for practical clustering tools. Future work might focus on optimizing computational efficiency and enhancing robustness for diverse and noisy datasets.

\section*{Acknowledgements}
This project was supported by the brazilian Ministry of Science, Technology and Innovations, with resources from Law nº 8,248, of October 23, 1991, within the scope of PPI-SOFTEX, coordinated by Softex and published Arquitetura Cognitiva (Phase 3), DOU 01245.003479/2024 -10.

\bibliographystyle{latex/splncs04}
\bibliography{bib}

\end{document}